\documentclass{article}

\usepackage{microtype}
\usepackage{graphicx}
\usepackage{subfigure}
\usepackage{booktabs} 
\usepackage{hyperref}

\usepackage[preprint]{icml2021}
\usepackage{natbib}
\usepackage{multirow}
\usepackage{enumitem}
\usepackage{paralist}
\newcommand{\kfx}{\mathbf{k_{fs}}}
\newcommand{\kfxp}{\mathbf{k_{fs'}}}

\usepackage{macros}
\icmltitlerunning{Tighter Bounds on the LML of GPR Using CG}

\begin{document}
\onecolumn

\icmltitle{Tighter Bounds on the Log Marginal Likelihood 
of Gaussian Process Regression Using Conjugate Gradients}


\icmlsetsymbol{equal}{*}

\begin{icmlauthorlist}
\icmlauthor{Artem Artemev}{equal,imp,sm}
\icmlauthor{David R.~Burt}{equal,cam}
\icmlauthor{Mark van der Wilk}{imp}
\end{icmlauthorlist}

\icmlaffiliation{imp}{Department of Computing, Imperial College London, London, United Kingdom}
\icmlaffiliation{sm}{Secondmind.ai, Cambridge, United Kingdom}
\icmlaffiliation{cam}{Department of Engineering, University of Cambridge, Cambridge, United Kingdom}

\icmlcorrespondingauthor{David R.~Burt}{drb62@cam.ac.uk}

\icmlkeywords{Gaussian processes, Scalable inference}
\vskip 0.3in

\printAffiliationsAndNotice{\icmlEqualContribution}

\begin{abstract}
We propose a lower bound on the log marginal likelihood of Gaussian process regression models that can be computed without matrix factorisation of the full kernel matrix. We show that approximate maximum likelihood learning of model parameters by maximising our lower bound retains many of the sparse variational approach benefits while reducing the bias introduced into parameter learning. The basis of our bound is a more careful analysis of the log-determinant term appearing in the log marginal likelihood, as well as using the method of conjugate gradients to derive tight lower bounds on the term involving a quadratic form. Our approach is a step forward in unifying methods relying on lower bound maximisation (e.g. variational methods) and iterative approaches based on conjugate gradients for training Gaussian processes. In experiments, we show improved predictive performance with our model for a comparable amount of training time compared to other conjugate gradient based approaches.
\end{abstract}

\section{Introduction} 
 Scaling models involving Gaussian process priors to large datasets is an important and well-researched problem in Bayesian statistics and machine learning. In order for a method to succeed in this task it should provide high-quality approximations to 1)~the posterior mean and (co-)variance over functions, 2)~the log marginal likelihood (LML). While only the former is needed for making predictions, the latter is a useful tool for model selection; when the kernel is differentiable with respect to hyperparameters, the LML or an approximation to it can be optimised using gradient based methods in order to automatically select model hyperparameters.
 
We derive a lower bound on the LML of regression with a Gaussian process prior and a Gaussian likelihood. We refer to this bound as CGLB. Parameter learning with CGLB combines many of the strengths of sparse variational Gaussian process regression (SGPR) \citep{titsias2009variational} with the strengths of conjugate gradient (CG) methods \citep{gibbs1997efficient} for GP inference. We use an improved bound on the log-determinant of the covariance matrix as well as CG to tighten the SGPR evidence lower bound (ELBO). We show empirically that this leads to less bias in parameter selection than maximisation of the ELBO. This reduced bias leads to improved performance on several benchmark tasks involving large datasets.

\Citet{wang2019exact} showed conjugate gradient based approaches to GP inference perform excellently on many regression tasks and argued that these methods should be considered `exact'. The latter claim has two caveats. First, gradient estimates provided by these methods are biased if conjugate gradients is stopped too early; while this bias can be reduced by running more iterations of CG, this comes at an additional computational cost. Second, the estimates of the LML are stochastic. The variance of the estimator provided can be reduced at the cost of needing to solve more systems of equations. Following \citet{davies2016thesis}, we refer to these approaches as `Iterative GPs', to contrast them with implementations using deterministic, direct methods for computing the LML.  

Building on work in \citet{gibbs1997efficient} and \citet{davies2016thesis} we derive a stopping criterion for our application of CG that ensures more iterations of CG would not improve the bound on the LML significantly. We also exploit old solutions to CG from previous iterations of hyperparameter learning. This approach allows us to often run zero or one steps of CG per iteration of hyperparameter learning without significantly impacting the approximate LML, even on large datasets.  Using a similar approach to \citet{gardner2018gpytorch,wang2019exact} and \citet{meanti2020kernel} CGLB can be implemented with memory complexity that is linear in the number of training examples, $n$, by splitting up (and parallelising) matrix-vector products \citep{charlier2020kernel}.

We empirically show that the combination of a deterministic objective function and reduced number of CG steps per iteration of hyperparameter optimisation leads to improved stability and performance when performing model selection (for comparable computational times) compared to existing Iterative GPs. 

\section{Background}\label{sec:background}

In this section, we review Gaussian process regression with a Gaussian likelihood. We then discuss the conjugate gradient method for solving linear systems, and how this can be applied to scaling Gaussian process regression. We conclude the section with a discussion of sparse variational inference as a method for scalable approximate inference in GP regression models.

\subsection{Gaussian Process Regression}
We assume a dataset has been observed and denote it by $\mathcal{D}=\{(x_
i,y_i)\}_{i=1}^n$ with $x_i \in \mathcal{X}$ where $\mathcal{X}$ denotes the set of possible observed features and $y_i \in \mathbb{R}$. Let $\ydat \in \mathbb{R}^n$ denote the vector formed by concatenating the $y_i$ and $\xdat \in \mathcal{X}^n$ denote the tuple $(x_i)_{i=1}^n$.

We take a Bayesian approach with prior $f\sim \mathcal{GP}(0, k)$, i.e.~$f$ is a Gaussian process with zero mean and covariance function $k$, and likelihood $Y|f(\xdat) \sim \mathcal{N}(f(\xdat), \sigma^2 \bfI)$, where $Y$ is an $\R^n$-valued random variable and by an abuse of notation we use $f(\xdat)\in \R^n$ to denote the $\R^n$-valued random variable formed by indexing $f$ at each $x_i$. Inference involves computing the distribution of $f|(Y=\ydat)$.
This is again a Gaussian process with mean and covariance,
\begin{align*}
    \tilde{\mu}(x) = \kfx \inv{\nKmat}\ydat \text{,\quad} \tilde{k}(x,x') = k(x,x') - \kfx\transpose \inv{\nKmat}\kfxp,
\end{align*}
where $\nKmat = \Kff + \noisevar\bfI$, $\Kff$ is the $n \times n$ matrix with entries $(\Kff)_{ij}=k(x_i,x_j)$, and $\kfx, \kfxp \in \R^n$ take values $(\kfx)_i = k(x_i,x)$ and $(\kfxp)_i = k(x_i,x')$. 

We assume $k$ is parameterised, and denote these hyperparameters together with $\sigma^2$ as $\theta$. The choice of $\theta$ has a significant impact on the generalisation properties of the posterior \citep[Chapter 5]{rasmussen2006gaussian}.

Type-II maximum likelihood is a heuristic that automates selection of $\theta$ by maximising the LML, defined as the log density of the prior probability distribution over $Y$ evaluated at $Y=\ydat$, with respect to the hyperparameters $\theta$. For this model the LML is
\begin{equation}
    \log p_Y(\ydat; \theta) = c - \frac{1}{2}\underbrace{\ydat\transpose \inv{\nKmat} \ydat}_{\text{quad. term}} - \frac{1}{2} \underbrace{\logdet{\nKmat}}_{\text{log-det. term}}. \label{eqn:gpr-lml}
\end{equation}
with $c=-\frac{n}{2}\log 2\pi$. Here and elsewhere, we suppress the dependence of kernel matrices on $\theta$. Common implementations of Gaussian process inference and hyperparameter selection rely on a Cholesky factorisation of $\nKmat$ in order to evaluate $\log |\nKmat|$ and $\inv{\nKmat}\ydat$. The Cholesky decomposition is usually implemented in a way that requires roughly $n^3/3$ floating point operations and stores a matrix with $n(n+1)/2$ distinct entries in memory. This can be a prohibitive cost for regression problems with many observations.

\subsection{Conjugate Gradients}
The conjugate gradient algorithm \citep{hestenes1952methods} is a method for solving systems of equations using only matrix-vector multiplication and elementary vector operations. Given a symmetric positive definite matrix $\nKmat$ and a vector $\ydat$ the goal of conjugate gradients is to find an vector $\vvec$ satisfying $\nKmat\vvec=\ydat$. Starting with an initial guess for $\vvec$, each iteration of conjugate gradient chooses a search direction and updates the current guess for $\vvec$ by adding a vector in the chosen direction.  In exact arithmetic, CG is guaranteed to solve an $n\times n$ system of equations in $n$ iterations, each of which involves a $\Theta(n^2)$ matrix-vector multiplication, viewed as an iterative algorithm CG has strong guarantees on the rate at which the error decreases, at least for well-conditioned matrices \citep[Section 10.2.3]{Hackbusch1994}. The convergence of conjugate gradient is often practically assessed by examining the residual $\bfr=\ydat-\nKmat\vvec$, which can be evaluated with a matrix-vector multiplication. If the residual is $0$, then the algorithm has converged. Often CG is stopped when the residual has a sufficiently small Euclidean norm. 

\subsection{Gaussian Process Regression with Conjugate Gradient Methods}

Conjugate gradients has been suggested as a method to directly approximate the gradient of \cref{eqn:gpr-lml} \citep{gibbs1997efficient}. An obstacle to this approach is the evaluation of the gradient of the log-determinant, $\frac{\partial}{\partial \theta_i} \log |\nKmat| = \mathrm{tr}(\inv{\nKmat}\frac{\partial\nKmat}{\partial \theta_i})$.  Evaluating this trace directly is computationally expensive, as it requires solving $n$, $n\times n$ systems of linear equations. Hutchinson's trace estimator \citep{hutchinson1989stochastic} provides a stochastic estimate of this gradient. The estimator is formed by first noting that for any $\R^n$-valued random variable $\bfp$ such that $\Exp{\bfp}{\bfp\bfp\transpose}=\bfI$,  $\mathrm{tr}(\inv{\nKmat}\frac{\partial\nKmat}{\partial \theta_i})=\Exp{\bfp}{\bfp\transpose\inv{\nKmat}\frac{\partial\nKmat}{\partial \theta_i}\bfp}$. The expectation can be estimated with Monte Carlo, using CG to approximate $\inv{\nKmat}\bfp_i$, where $\bfp_i$ is a sample of $\bfp$.

This results in a biased, stochastic estimate of the gradient. The bias in this estimator can be decreased, and practically removed, at the cost of increasing the number of iterations of CG. The variance can be reduced by increasing the number of $\bfp_i$ samples, at the cost of needing to solve more systems of equation. The variance of this estimator for various distributions of $\bfp$ as well as high probability bounds on the relative error are known \citep{avron2011randomized}.

In cases when the objective function itself is of interest, for example if model comparison is performed with discrete hyperparameters, approximations to the LML using CG and related ideas have also been proposed \citep{ubaru2017fast}. Iterative GPs have been shown to be highly scalable with modern computational architectures \citep{gardner2018gpytorch,wang2019exact}.

\subsection{Gaussian Process Regression with Sparse Methods}
An alternative approach, which avoids the computation of $\nKmat$ entirely, relies on sparsity assumptions in the data-domain that lead to a low-rank approximation of $\Kff$, \citep[e.g.][]{williams2001using, snelson2005sparse}. This approach reduces the computational requirement to $\mcO(nm^2)$ where $m$ is a parameter that controls the rank of the approximation to $\Kff$. \Citet{titsias2009variational} proposed an interpretation of sparse methods as variational inference with a structured family of posterior distributions. This framework defines an evidence lower bound (ELBO), $L(\bfy;\theta)\leq \log p(\ydat; \theta)$, where
\begin{align}
L(\bfy;\theta) &= c -\frac{1}{2} \underbrace{\ydat\transpose \nQmat^{-1}\ydat}_{\substack{\text{bound on} \\ \text{quad. term}}} - \frac{1}{2}\underbrace{\left(\log |\nQmat|+\frac{\mathrm{tr}(\Kff - \Qff)}{\sigma^2}\right)}_{\text{bound on log-det. term}} \label{eqn:elbo}
\end{align}
with $\nQmat=\Qff + \sigma^2\bfI$, with $\Qff=\Kuf\transpose \Kuu\Kuf$, $\Kuu$ an $m\times m$ matrix with $(\Kuu)_{ij}=k(z_i,z_j)$, $\Kuf$ an $m \times n$ matrix with $(\Kuf)_{ij}=k(z_i,x_j)$, and the $z_i\in \mathcal{X}$ are variational parameters. The ELBO can be evaluated in $O(nm^2)$, and is frequently jointly maximised with respect to variational parameters $\{z_i\}_{i=1}^m$ and hyperparameters $\theta$ as a form of approximate maximum likelihood learning. The variational posterior is then used for prediction. This is a Gaussian process, with mean and covariance,
\begin{align}
   \widehat{m}(x) &= \kux\transpose\inv{\Kuu}\Kuf\inv{\nQmat}\ydat \text{\quad and \quad} \label{eqn:mean-sgpr}\\
   \hat{k}(x,x') &= k(x,x') - \kux\transpose\inv{\Kuu}\Kuf \inv{\nQmat}\Kuf\inv{\Kuu}\kuxp.\label{eqn:cov-sgpr}
\end{align}
where $\kux,\kuxp\in \R^m$ have entries $\kux_i=k(\bfz_i,x)$ and $\kuxp_i=k(\bfz_i,x')$.
When the number of training examples is large, the kernel is sufficiently smooth, and the data is not too spread out $\log p_Y(\bfy;\theta) - L(\bfy;\theta)$ is small for some $m \ll n$ \citep{JMLR:v21:19-1015}, in which case one expects that similar hyperparameters would be selected by maximising the ELBO as would be selected by maximising the LML. However, certain settings of $\theta$, for example those for which $\sigma^2$ is very small, lead to large discrepancies between the LML and ELBO, which results in significant bias in parameter selection and underfitting \citep{bauer2016understanding}. 

\section{Lower Bounds on the Log Marginal Likelihood}

In this section we present our lower bound on the log marginal likelihood that can be computed in $O(nm^2+(t+1)n^2)$, where $t$ is the number of steps of CG run. We also present a mean function that can be used for prediction in conjunction this bound. We additionally derive lower bounds on the LML that are tighter than the evidence lower bound \cref{eqn:elbo} and can be computed in $O(nm^2)$.  While we focus on the CG version of our bounds, which we refer to as CGLB, these bounds may be of independent interest.

\subsection{Conjugate Gradient Lower Bound}
We now state the lower bound, we refere to as CGLB:
\begin{lem}\label{lem:lower-bound-lml}
Let $\nKmat$ as in \cref{eqn:gpr-lml}, $\nQmat$ as in \cref{eqn:elbo}, $c=-\frac{n}{2}\log 2\pi$ and for any $\vvec \in \R^n$
\begin{align}
    \log p(y; \theta) \geq c-\frac{1}{2} \left(\resid \transpose \inv{\nQmat} \resid + 2\ydat\transpose \vvec - \vvec\transpose \nKmat \vvec\right) -\frac{1}{2}\left(\logdet{\nQmat} + n \log \left(1 + \frac{\Tr(\nKmat - \nQmat)}{n\sigma^2}\right)\right),\label{eqn:lml-bound}
\end{align}
where $\resid = \ydat - \nKmat\vvec$.
\end{lem}

The right hand side of \cref{eqn:lml-bound} can be maximised with respect to $\{\theta, \vvec, \{z_i\}_{i=1}^m\}$ for parameter learning. For fixed $\vvec$, the right hand side can be computed in $O(n^2+nm^2)$ using similar computations to \cref{eqn:elbo} but with an additional $n\times n$ matrix-vector product to compute $\nKmat\vvec$. If $\vvec=\inv{\nKmat}\ydat$, then $\bfr=0$, and the first term in the bound becomes $-\frac{1}{2}\ydat\inv{\nKmat}\ydat$. This choice of $\vvec$ maximises the lower bound (CGLB) for any choice of $\theta,\{z_i\}_{i=1}^m$, and is independent of $\{z_i\}_{i=1}^m$, and can be approximated by running conjugate gradients on the system of equation $\nKmat \vvec=\ydat$. 

In order to derive this lower bound on the log marginal likelihood, it suffices to upper bound the quadratic term and the log-determinant term from \cref{eqn:gpr-lml}. 

\subsection{Bounds on the Log-Determinant Term}

Unlike previous Iterative GPs \citep{gibbs1997efficient, gardner2018gpytorch}, we aim for a deterministic estimate of the LML and its gradient. A simple approach is to combine estimates of the term $\ydat\transpose \inv{\nKmat}\ydat$ based on CG with the bound $\log |\nKmat| \leq \log |\nQmat| + \frac{1}{\sigma^2}\Tr(\nKmat- \nQmat)$ from SGPR. This bound is tight when $\Tr(\nKmat-\nQmat) \approx 0$, but is loose otherwise. We derive a tighter bound on $\log |\nKmat|$ using properties of $\nQmat$. 

We recall several matrix properties. 

\begin{prop}[{\citealp[page 51]{horn2012matrix}}]\label{prop:det-trace}
Let $A \in \R^{k \times k}$ and $\{\lambda_i\}_{i=1}^k$ denote the eigenvalues of $A$ (counted with multiplicity). Then $|A| = \prod_{i=1}^k \lambda_i$ and $\Tr(A) = \sum_{i=1}^k \lambda_i$.
\end{prop}

We say a symmetric matrix $\bfA \in \R^{k\times k}$ is \emph{positive semi-definite} (PSD) if for all $\mathbf{z} \in \R^{k}, \bfz\transpose\bfA\bfz \geq 0$. For any PSD matrix, we have $\lambda_i \geq 0$ where $\lambda_i$ denotes an eigenvalue of $\bfA$.

\begin{prop}[{\citealp[page 495]{horn2012matrix}}]\label{prop:schur-complement-psd}
Let $H$ be a symmetric real matrix with $H =\left[\begin{smallmatrix}
     A& B \\
     B\transpose& C
\end{smallmatrix}\right]$
with $A$ non-singular. Then $H$ is PSD if and only if $A$ and $C-B\transpose A^{-1} B$ are both PSD.
\end{prop}

Applying this lemma to the block matrix $H =\left[ \begin{smallmatrix}
     \Kuu & \Kuf \\
     \Kuf\transpose& \Kff
\end{smallmatrix}\right]$, which is PSD since the kernel is PSD, leads to the conclusion $\Kff -\Qff$ is PSD.

\begin{prop}[{\citealp[Corollary 7.7.4]{horn2012matrix}}]\label{prop:psd-ordering-props}
Let $A_1, A_2 \in \R^{k\times k}$ such that $A_1-A_2$ is PSD. Let $\lambda_i(A_j), 1\leq i\leq k, j \in \{1,2\}$ denote the $i$\textsuperscript{th} largest eigenvalue of $A_j$. Then, 
\begin{compactenum}
    \item If $A_1$ and $A_2$ are invertible, $A_2^{-1} - A_1^{-1}$ is PSD,
    \item $\lambda_i(A_1) \geq \lambda_i(A_2)$ for $1\leq i \leq k$.
\end{compactenum}
\end{prop}

We now derive the SGPR bound $\log |\nKmat| \leq \log |\nQmat| + \frac{1}{\sigma^2}\Tr(\Kff-\Qff)$; the bound on the log-determinant term used in \cref{lem:lower-bound-lml} is derived by a modification of this argument and the arithmetic-geometric mean inequality.

Let $\lambda_i$ denote the eigenvalues of $\nKmat$ and $\ell_i$ denote the eigenvalues of $\nQmat$, both sorted in descending order. Adding zero and using \cref{prop:det-trace}, 
\begin{align}
    \logdet{\nKmat} = \logdet{\nQmat} + \logdet{\nKmat} -  \logdet{\nQmat}
     = \logdet{\nQmat} + \sum_{i=1}^n \log\left(1+ \frac{\lambda_i-\ell_i}{\ell_i}\right).
\end{align}
Since $\nQmat - \noisevar \bfI=\Qff$ is PSD, we may apply \cref{prop:psd-ordering-props} to conclude,
\begin{align}
     \sum_{i=1}^n \log\left(1+ \frac{\lambda_i-\ell_i}{\ell_i}\right) \leq  \sum_{i=1}^n \log\left(1+ \frac{\lambda_i-\ell_i}{\noisevar}\right) \label{eqn:partial-log-det}.
\end{align}
Applying the inequality $\log(1+x) \leq x$ for $x>-1$ to each summand,
\begin{align}
  \sum_{i=1}^n \log\left(1+ \frac{\lambda_i-\ell_i}{\noisevar}\right)& \leq \frac{1}{\noisevar}\left(\sum_{i=1}^n \lambda_i-\sum_{i=1}^n \ell_i\right) \label{eqn:log-bound}\\
    & =  \frac{1}{\noisevar}\Tr(\nKmat - \nQmat) \label{eqn:titsias-log-det}.
\end{align}

Note that the bound $\log(1+x) \leq x$ is only tight for $x=0$. Appying this elementwise can lead to loose bounds in cases when $\Tr(\Kff-\Qff)$ is large. We would like to replace the inequality in \cref{eqn:log-bound} with a tighter inequality. We recall the arithmetic-geometric mean inequality.

\begin{prop}[AM-GM inequality; {\citealp[page 559]{horn2012matrix}}]\label{prop:am-gm}
Let $a_1, \dotsc, a_n\geq 0$, then 
\begin{equation}
    \left(\prod_{i=1}^n a_i\right)^{1/n} \leq \frac{1}{n} \sum_{i=1}^n a_i.
\end{equation}
\end{prop}

We now state the bound on the log-determinant.
\begin{lem}\label{lem:logdet-bound}
For  $\nKmat,\nQmat \in \R^{n \times n}$ PSD such that $\nKmat-\nQmat$ is PSD and $\nQmat-\sigma^2 \bfI$ is PSD,
\begin{align}
    \logdet{\nKmat} &\leq \logdet{\nQmat} + n \log \left(1 + \frac{\Tr(\nKmat - \nQmat)}{n\sigma^2}\right)\label{eqn:logdet-bound}\\
    &\leq \logdet{\nQmat} + \frac{1}{\noisevar}\Tr(\nKmat - \nQmat).\label{eqn:log-det-sgpr}
\end{align}
\end{lem}
A related result to \cref{lem:logdet-bound} was proven concurrently to this work in \citep{vakili2020information}, and used in deriving a bound on the information gain of Gaussian process models. 
\begin{proof}[Proof of \cref{lem:logdet-bound}]

By \cref{prop:schur-complement-psd} and \cref{prop:psd-ordering-props}, $\lambda_i-\ell_i$ is non-negative for all $i$. We can therefore apply  \cref{prop:am-gm} to the left hand side of \cref{eqn:log-bound},
\begin{align*}
   \sum_{i=1}^n \log\left(1+ \frac{\lambda_i-\ell_i}{\noisevar}\right)
    \leq n \log \left( 1 + \frac{\sum_{i=1}^n \lambda_i - \ell_i }{n\sigma^2}\right) 
    = n \log \left(1 + \frac{\Tr(\nKmat - \nQmat)}{n\sigma^2}\right). 
\end{align*}

\Cref{eqn:log-det-sgpr} follows from $\log(1+x) \leq x$ for all $x>-1$.

\end{proof}

Like the SGPR bound on the log-determinant term, we only need to compute $\log |\nQmat|$ and $\Tr(\Kff-\Qff)$. We could use \cref{lem:logdet-bound} together with the bound on the quadratic form from SGPR, giving a lower bound on the LML that can be computed in $O(nm^2)$, with only minor modifications to standard implementations of SGPR. 

We show in the appendix that this bound can be further improved using a similar approach, by phrasing the problem of upper bounding the log-determinant of $\nKmat$ as a constrained convex optimisation problem, subject to the constraints that $\nKmat - \nQmat$ is PSD, and $\Tr(\nKmat)$ as well as the eigenvalues of $\nQmat$ are known. For simplicity, we run experiments using \cref{lem:logdet-bound}.

\subsection{Bounds on the Quadratic Term}
We now turn our attention to an upper bound on the quadratic form $\ydat\transpose \inv{\nKmat} \ydat$. In order to combine the benefits of SGPR and iterative methods derive an upper bound on $\ydat\transpose \inv{\nKmat}\ydat$ that is tight if either: 1) $\Qff \approx \Kff$ 2) we are able to find a $\vvec \in \R^n$ such that $\vvec \approx \inv{\nKmat} \ydat$.

\begin{lem}\label{prop:bound-quad-term}
 Let $\nKmat, \nQmat \in \R^{k \times k}$ PD such that $\nKmat-\nQmat$ is PSD. Then, for any $\vvec, \ydat \in \R^{k}$,
\begin{align}
        2\ydat\transpose \vvec - \vvec\transpose \nKmat \vvec \leq \ydat\transpose \inv{\nKmat} \ydat \leq  \resid \transpose \inv{\nQmat} \resid + 2\ydat\transpose \vvec - \vvec\transpose \nKmat \vvec,
\end{align}
where $\resid = \ydat - \nKmat \vvec$.
\end{lem}
\begin{rem}
The lower bound is standard, and yields the popular interpretation of CG as optimisation of a quadratic function, \citep[e.g.~][Section 9.1.1]{Hackbusch1994}
\end{rem}
Choosing $\vvec = \zerovec$ results in the upper bound $\ydat\transpose \inv{\nKmat} \ydat \leq \ydat\transpose \inv{\nQmat} \ydat$, which corresponds to the quadratic term in the SGPR bound. If we have $m=0$ so that $\nQmat=\sigma^2\bfI$, after some rearranging we recover the upper bound considered in \citet[eqn.~ 50]{gibbs1997efficient} for monitoring the convergence of CG.

\begin{proof}[Proof of \cref{prop:bound-quad-term}]
We begin by expanding out the quadratic form,
\begin{align}
    \ydat\transpose \inv{\nKmat} \ydat = (\resid + \nKmat \vvec)\transpose \inv{\nKmat} (\resid + \nKmat \vvec) 
     = \resid\transpose\inv{\nKmat}\resid +2\resid\transpose\vvec +\vvec\transpose\nKmat\vvec \label{eqn:partial-quadterm}.
\end{align}
Let $\bfw = \inv{\nKmat}\ydat$. Then  $\resid\transpose\inv{\nKmat}\resid = \bfw \transpose \nKmat \bfw \geq 0$. This proves the lower bounds in \cref{prop:bound-quad-term}.

Adding $0$ to the term involving an inverse,  
\begin{align}
    \resid\transpose\inv{\nKmat}\resid = \resid\transpose\inv{\nQmat}\resid - \resid\transpose(\inv{\nQmat}-\inv{\nKmat})\resid.
\end{align}
From \cref{prop:psd-ordering-props} and the assumption that $\nKmat - \nQmat$ is PSD $\inv{\nQmat}-\inv{\nKmat}$ is PSD. Hence, $\resid\transpose\inv{\nKmat}\resid \leq \resid\transpose\inv{\nQmat}\resid. $
\end{proof}

\subsection{Approximate Predictive Posterior}\label{sec:predictive}

After selecting model hyperparmeters, we must compute a mean and (co-)variance at test points. We use the same covariance calculation as in SGPR (\cref{eqn:cov-sgpr}). For the mean, we would like to recover the exact GP mean if $\nKmat=\nQmat$ or $\vvec=\nKmat^{-1}\bfy$, as in both cases our bound estimates $\ydat\transpose\inv{\nKmat}\ydat$ exactly. We propose the estimator
\begin{equation}
    m(x) = \kfx\transpose\vvec +\kux\transpose\Kuu^{-1}\Kuf\inv{\nQmat}(\ydat-\nKmat \vvec), \label{eqn:cglb-mean}
\end{equation} 
where $\kux$ is as in \cref{eqn:cov-sgpr} and $\kfx \in \R^n$ with $\kfx_i = k(x_i,x)$. The first term is the estimate given by running conjugate gradient and replacing $\inv{\nKmat}\ydat$ with its approximation. The second term is a correction that is equivalent to the mean of SGPR, but replacing $\bfy$ with the residual of the CG computation. The mean predictor \cref{eqn:cglb-mean} is not sparse, but the covariance we use is induced in the same way as in the sparse variational framework. The CGLB predictive posterior closely resembles `decoupled' variational Gaussian process inference \citet{NIPS2017_f8da71e5}, with the inducing points used to determine the mean function $Z'=\{X,Z\}$ and the inducing points used to determine the covariance function $Z$.

A worthwhile question is whether CGLB has a variational interpretation, i.e.~whether the gap between CGLB and the LML is a Kullback-Leibler divergence between an approximate posterior and the posterior. Variationally formulated bounds would mean that the implicit approximate posterior provides an alternative predictive distribution to the  predictive distribution we describe in \cref{sec:predictive}.

\subsection{Implementation with Conjugate Gradient}\label{sec:cg-implementation}
The vector $\vvec \in \R^{n\times n}$ is an auxiliary parameter in our lower bound on the LML. From \cref{prop:bound-quad-term}, for fixed $\theta$, CGLB is maximized by $\vvec = \inv{\nKmat}\ydat$. A simple approach is to treat $\vvec$ as an additional parameter, and optimise the bound jointly with respect to $\{\theta, \vvec, \{z_i\}_{i=1}^m\}$. However, this introduces $n$ additional dimensions to the optimisation problem, and we find it leads to slow parameter learning.

\begin{figure}[t!]
    \centering
    \includegraphics[width=0.6\textwidth]{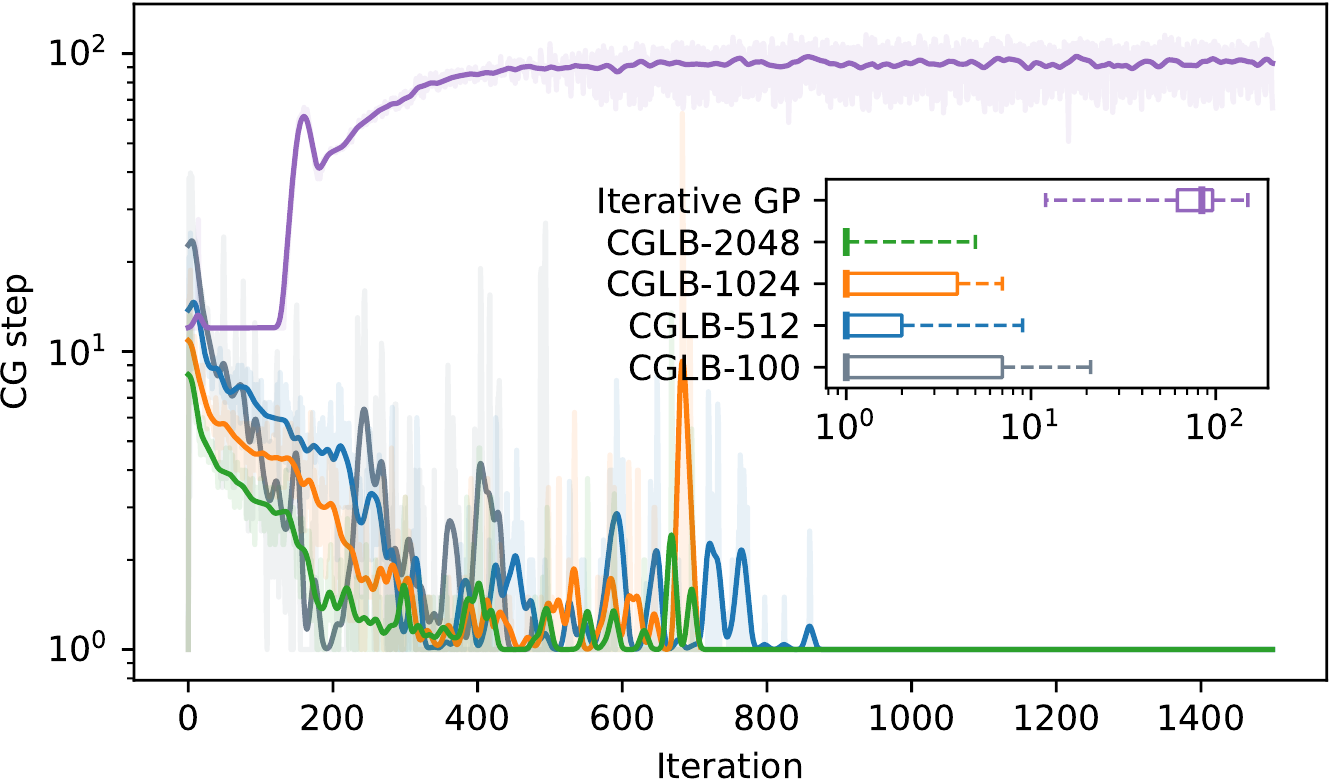}
    \vspace{-.2cm}
    \caption{The number of CG iterations spent in CGLB and Iterative GP to achieve a pre-set residual error (note the stopping criteria for the methods is different) on the \texttt{protein} dataset. For CGLB this number goes to zero (the plot uses $\log(x + 1)$ scaling for the CG step axis), and CGLB reuses the vector $v$ in the subsequent iterations. The figure shows the average number of steps (shaded lines) spent per function evaluation during optimisation for five experiments on a different data splits; solid lines are the smoothing applied to the average number of steps for each model. The inner box-whisker plot depicts IQRs of CG steps that CGLB and Iterative GP models ran throughout 1500 iterations. The whiskers set to 95 percentile. The initial flat line for Iterative GP at 10 iterations is due to a hard-coded constraint in the GPytorch code.}
    \label{fig:cg-steps}
\end{figure}

We instead select $\vvec$ by running conjugate gradient on the system of equations $\nKmat \vvec = \ydat$ each time we evaluate the lower bound on the LML. We make several design choices when running conjugate gradients to improve and assess convergence of the approximation.
\paragraph{Preconditioner}
We use $\nQmat$ as a preconditioner. This preconditioner has been used previously in kernel methods \citep{cutajar2016preconditioning}. In our case, this has computational advantages: we can reuse some of the calculations used in bounding the log-determinant. 
\paragraph{Initializing CG}
We initialize CG using the value of $\vvec$ found in the previous evaluation of the lower bound. As the optimiser often evaluates similar settings of kernel hyperparameters in sequence, as long as $\sigma^2$ is not very close to zero, similar parameter settings generally result in similar optimal values for $\vvec$ this is often a good guess for the the solution to $\inv{\nKmat}\ydat$. We show in \cref{sec:experiments} for many datasets, after the first handful of optimisation steps the old solution for $\vvec$ is good enough
and no iterations of CG are needed during most steps of parameter optimisation.
\paragraph{Stopping Criterion}
We monitor upper and lower bounds on $\ydat\transpose \inv{\nKmat} \ydat$ in order to decide when to terminate CG, as advocated in \citet{gibbs1997efficient}. Subtracting the upper and lower bounds in \cref{prop:bound-quad-term}, we derive the stopping criteria $\resid\transpose \inv{\nQmat}\resid \leq 2\epsilon$. This ensures that the slack in our bound introduced from not exactly computing the quadratic term is at most $\epsilon$. Because of our choice of preconditioner, $\resid\transpose \inv{\nQmat}\resid$ is computed in every iteration, so this stopping criterion requires only a single extra matrix-vector multiplication with the preconditioner in the final iteration. 

Although stopping criteria based on $\|\bfr\|_2$ are a popular heuristic and often practical, they can lead to significant biases in parameter selection in this application. In particular, the bias in the lower bound (the gradient of which corresponds to the gradient estimate used in \citet{gardner2018gpytorch}) in \cref{prop:bound-quad-term} is $\bfr\transpose\inv{\nKmat}\bfr$. In the worst-case the bias introduced into the LML can be as large as $\|\bfr\|^2_2/(2\sigma^2)$. A similar argument shows that in the worst case and when $m=0$, the upper bound in \cref{prop:bound-quad-term} could have the same bias in the opposite direction. By instead ensuring $\bfr\transpose\inv{\nQmat}\bfr\leq 2\epsilon$, we ensure the bias we introduce through this term in CGLB is uniformly bounded by $\epsilon$ over all parameter settings.  We therefore expect this bias to have a smaller impact on the estimation of hyperparameters, in particular $\noisevar$.

\begin{figure*}[h!]
    \centering
    \includegraphics[width=0.87\textwidth]{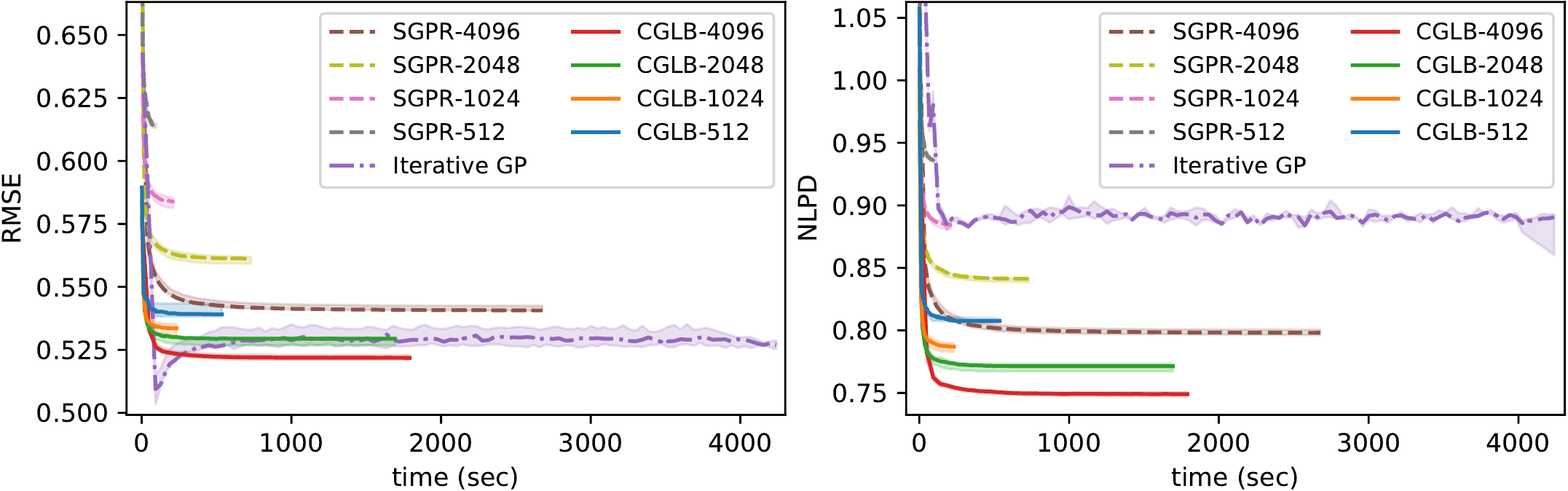}
    \caption{Test root mean square error (RMSE) and negative log predictive density (NLPD) metrics of CGGP, SGPR and Iterative GP models computed on the \texttt{protein} dataset. The shaded area is IQR region, and the line is median over five experiment trials with different dataset splits.}
    \label{fig:protein-rmse-nlpd}
\end{figure*}

\subsection{Selecting Optimization Parameters with CGLB versus Iterative Methods}
An advantage of our approach compared to existing iterative methods is that the quality of design and optimization choices can be assessed directly through the approximation to the log marginal likelihood, as is done in Cholesky-based implementations of GPR and SGPR. Good choices of the parameters $\vvec, \{z_i\}_{i=1}^m$ result in tighter lower bounds on the log marginal likelihood. 

In contrast, when using Iterative GP any bias introduced due to an insufficient number of iterations of CG may lead to either over or under estimation of the log marginal likelihood. In this case, changes to optimization parameters that increase the estimated log marginal likelihood could be indicative of additional bias instead of improved hyperparameter selection. Additionally, as the estimate is stochastic, without many evaluations of the LML or many vectors used in Hutchinson's trace estimator, it may be difficult to tell which of two models is preferred by the estimate of the LML.

While methods considered in \citet{gibbs1997efficient, ubaru2017fast, gardner2018gpytorch, wang2019exact} are capable of having less bias when estimating the LML than our method (at least for fixed $m, \theta$), any bias introduced is harder to assess. In the next section, we find empirically this makes achieving good performance with these methods more challenging.

\section{Experiments}\label{sec:experiments}

We now compare hyperparameter selection with CGLB (\cref{lem:lower-bound-lml}), with sparse methods (SGPR) as well as the conjugate gradient method considered in \citet{wang2019exact} (Iterative GP). We would like to determine which method finds better hyperparamters, and how these hyperparameters effect predictive performacen. We consider several regression UCI datasets, and compare the methods on the basis of root mean square error (RMSE) and negative log predictive density (NLPD) on held-out data to assess predictive performance. In order to assess the quality of the hyperparameters found by the methods, for datasets where it is computationally feasible, we compare the LML of the model with the hyperparameters selected by each method, computed directly with Cholesky decomposition.

\subsection{Data Preparation}
We randomly split each dataset into a training set consisting of 2/3 of examples, and a test set consisting of the remaining 1/3. We run 5 seeds in all experiments, with each seed corresponding to a different random split of the dataset. Each input dimension is normalised to have mean $0$ and variance $1$ within the training set. Similarly, the training outputs are normalised to have $0$ mean and variance $1$. We apply the same normalisation to test data when making predictions, using the statistics computed on the training data. All metrics reported are on the standardised data, so that we would expect a model predicting a constant mean function to have RMSE near $1.0$. Dataset are downloaded using the Bayesian benchmarks package \citep{Salimbeni}.

\subsection{Model Class and Initialisation of Parameters}

We run experiments with a Mat\'ern 3/2 kernel, with independently learned lengthscales along each input dimension (i.e.~automatic relevance detection). While in the derivations we have assume the prior mean is $0$, in experiments we take the prior mean to be a constant function that is learned as a hyperparameter. The changes to the predictive distribution and lower bound presented in the previous section to account for this are straightforward. This is the same experimental setup considered in \citet{wang2019exact}. 

All kernel lengthscales, the kernel variance and the likelihood variance are initialised at $1.0$. The prior mean is initialised at $0$. We use soft-plus constraints on the lengthscales and likelihood, lower bounding them at $1\mathrm{e}{-6}$ for SGPR and CGLB and $1\mathrm{e}{-4}$ for iterative GP, as we found this was necessary for numerical stability. Experiments are run using double precision.

\subsection{Training Procedure}
For CGLB and SGPR we vary the number of inducing points between 512 and 4096. We train CGLB and SGPR with the L-BFGS optimiser \citep{liu1989limited} for either 2000 steps or until the optimiser stops due to a small projected gradient norm or being unable to find a point that improves upon the current value during line-search. We initialise $\{z_i\}_{i=1}^m$ using the `Greedy' method advocated for in \citet{JMLR:v21:19-1015}, then optimise them jointly with hyperparameters. Alternatives to joint optimisation, such as reinitialisation of inducing points as suggested in \citet{JMLR:v21:19-1015} may lead to further improvements in the training procedure, though we do not explore those here. For CGLB we stop CG when the residual $\bfr$ satisfied $\frac{1}{2}\bfr\transpose\inv{\nQmat}\bfr \leq \epsilon=1.0$ when $\frac{1}{2}\bfr\transpose\inv{\nQmat}\bfr \leq \epsilon=1\mathrm{e}{-3}$ when selecting $\vvec$ for use in predictions. The criteria $\epsilon=1.0$ used during training ensures that the additional slack introduced into CGLB from estimating the quadratic term is at most $1.0$, regardless of the current value of $\theta$.

We follow the procedure described in \citet{wang2019exact} to train Iterative GP. We first pre-train the model on a subset of $10000$ observations with $10$ iterations of L-BFGS. Then we run $10$ iterations of Adam optimiser \citep{kingma2014adam} on the same subset, followed by $2000$ iterations of Adam with $0.1$ learning rate on the full dataset. For forming Hutchinson's trace estimation, $10$ vectors $\{\bfp_i\}_{i=1}^{10}$ are sampled. In order to run L-BFGS, the vectors $\{\bfp_i\}_{i=1}^{10}$ are fixed along each line search, as suggested in the documentation for GPytorch \citep{gardner2018gpytorch}.

For experiments with SGPR we use GPflow \citep{gpflow2017}. For small datasets ($n<20000$), we use a GPflow implementation of CGLB as we found this to be faster than our GPytorch implementation. However, for larger datasets we use GPytorch kernel abstractions with KeOps \citep{charlier2020kernel} to perform matrix-vector operations to reduce memory requirements. For Iterative GPs we use the GPytorch implementation \citep{gardner2018gpytorch}. Following \citet{wang2019exact}, we use the Lanczos variance estimator (without kernel interpolation) introduced in \citet{DBLP:conf/icml/PleissGWW18} for computing the predictive NLPD for Iterative GPs. We run all experiments on a single Tesla V100-32GB GPU.

\subsection{Results}

\begin{table*}[h!]
    \caption{Median LML, predictive NLPD and predictive RMSE over five datasets splits for Iterative GP, SGPR and CGLB. Cholesky subcolumns represent the same metrics evaluated by using Cholesky-based GPR implementation with hyperparameters found by Iterative GP, SGPR and CGLB models accordingly.\vspace{.2cm}}
    \label{tab:gpr-metrics}
    \centering
    \footnotesize
    \begin{tabular}{cccc|cc|cc}
    \toprule
    \multirow{2}{*}{} & \multirow{2}{*}{} 
    & \multicolumn{2}{c|}{LML} & \multicolumn{2}{c|}{NLPD} & \multicolumn{2}{c}{RMSE} \\
    & {} & \multicolumn{1}{c}{Approx} & \multicolumn{1}{c|}{Cholesky} & \multicolumn{1}{c}{Approx} & \multicolumn{1}{c|}{Cholesky} & \multicolumn{1}{c}{Approx} & \multicolumn{1}{c}{Cholesky} \\
    \midrule
    \multirow{4}{*}{\shortstack{\texttt{bike}\\n=17379, d=17}}
    & Iterative GP  & 30992.8      & 31319.1      & -2.016      & -3.257       & 0.014      &  0.020      \\
    & SGPR-4096     & 30502.5      & 32814.2      & -3.280      & -3.336       & 0.010      &  0.010      \\
    & CGLB-4096     & 37732.7      & \bf{42023.0} & \bf{-4.216} & -4.329       & 0.004      &  0.004      \\
    & CGLB-2048     & 34102.8      & 38936.7      & -3.811      & -3.972       & \bf{0.003} &  0.003      \\
    & CGLB-1024     & 30493.9      & 35351.8      & -3.403      & -3.615       & 0.005      &  0.005      \\
    \hline
    \multirow{4}{*}{\shortstack{\texttt{elevators}\\n=16599, d=18}}
    & Iterative GP & -4709.0      & -4705.1       & 0.407      & 0.384.       & \bf{0.353} & 0.353      \\
    & SGPR-4096    & -4675.3      & \bf{-4653.3}  & \bf{0.386} & 0.386        & 0.354      & 0.354      \\
    & CGLB-4096    & -4669.8      & -4659.1       & \bf{0.386} & 0.386        & 0.354      & 0.354      \\
    & CGLB-2048    & -4677.9      & -4656.4       & 0.387      & 0.387        & 0.355      & 0.355      \\
    & CGLB-1024    & -4712.0      & -4670.0       & 0.392      & 0.391        & 0.356      & 0.356      \\
    \hline
    \multirow{4}{*}{\shortstack{\texttt{poletele}\\n=15000, d=26}}
    & Iterative GP &  13552.5     & -7641.5        & -0.935      & 1.217       & 0.079      & 0.078      \\
    & SGPR-4096    &  9057.7      &  9624.0        & -1.172      & -1.180      & 0.078      & 0.078      \\
    & CGLB-4096    &  9377.1      &  \bf{9862.2}   & \bf{-1.201} & -1.203      & \bf{0.077} & 0.077      \\
    & CGLB-2048    &  8248.6      &  9248.0        & -1.126      & -1.145      & 0.080      & 0.080      \\
    & CGLB-1024    &  7250.7      &  8694.2        & -1.057      & -1.098      & 0.083      & 0.083      \\
    \hline
    \multirow{4}{*}{\shortstack{\texttt{kin40k}\\n=40000, d=8}}
    & Iterative GP & 23859.0      & ---       & -0.454      & ---      & 0.087      & --- \\
    & SGPR-4096    & 7486.0       & ---       & -0.705      & ---      & 0.107      & --- \\
    & CGLB-4096    & 12244.2      & ---       & \bf{-0.919} & ---      & \bf{0.086} & --- \\
    & CGLB-2048    & 10028.6      & ---       & -0.826      & ---      & 0.088      & --- \\
    & CGLB-1024    &  7260.0      & ---       & -0.714      & ---      & 0.093      & --- \\
    \hline
    \multirow{4}{*}{\shortstack{\texttt{protein}\\n=45730, d=9}}
    & Iterative GP & -25703.9      & ---       & 0.897      & ---      & 0.531      & --- \\
    & SGPR-4096    & -27714.6      & ---       & 0.798      & ---      & 0.541      & --- \\
    & CGLB-4096    & -26570.7      & ---       & \bf{0.749} & ---      & \bf{0.522} & --- \\
    & CGLB-2048    & -27442.0      & ---       & 0.771      & ---      & 0.529      & --- \\
    & CGLB-1024    & -28243.7      & ---       & 0.790      & ---      & 0.535      & --- \\
    \bottomrule

\end{tabular}
\end{table*}

In \cref{fig:cg-steps}, we show that CGLB typically only needs a single matrix-vector multiply with the kernel matrix, and reuses $\vvec$ in most training steps. Adding more inducing points, which results in a better preconditioner, decreases the number of steps of optimisation before we use zero CG steps in most subsequent parameter updates. Iterative GP requires a considerably greater number of CG steps, particularly in later training steps.

In \cref{fig:protein-rmse-nlpd}, we demonstrate the predictive performance of CGLB during training and compare it with Iterative GP and SGPR on the `protein' UCI dataset \citep{Dua:2019}. After splitting, the `protein' dataset has $29267$ training examples, $16463$ testing examples and input dimensionality $9$. The predictive RMSE of CGLB is either better or equal to Iterative GP and consistently outperforms SGPR. Moreover, CGLB shows improved predictive uncertainty estimations as measured by NLPD compared to Iterative GP and SGPR. For CGLB and SGPR, increasing the number of inducing points leads to large performance gains. SGPR is faster per iteration than CGLB for a fixed $m$; however, achieving comparable performance to CGLB requires many more inducing points with SGPR on several datasets (\texttt{bike}, \texttt{kin40k}, \texttt{protein}). 

Results for other datasets are shown in \cref{tab:gpr-metrics}. For small datasets where we can compute the log marginal likelihood using Cholesky-based methods, we see that CGLB often finds settings of hyperparameters with higher LML than competing methods, suggesting that the gains in predictive performance are attributable to improved hyperparameter selection. We also note that on some datasets, Iterative GP significantly overestimates the LML. We hypothesise this is due to the interaction between bias and optimisation that can occur when maximising an estimate that is not a lower bound. The supplementary material shows model performance over time for several more regression datasets.

For several datasets, we found that Iterative GP achieved good test performance as measured by RMSE after a handful of iterations, but that continuing to optimise the objective lead to a small dropoff in performance, e.g.~\cref{fig:protein-rmse-nlpd}. We initially thought this might be due to a high learning rate. However, in the supplement, we show that even if the learning rate for Adam is reduced from $0.1$ to $0.01$ the same phenomena occurs, but convergence is slower. In contrast, we do not see this behaviour for SGPR or CGLB. 

Further, we observed unusual behaviour of the Iterative GP method on some datasets when the positive constraint of likelihood noise is set to a low number. In Iterative GP experiments, we used the GPytorch default noise value $1\mathrm{e}{-4}$. We attempted to lower this constraint to $1\mathrm{e}{-6}$ as the noise variance on several datasets selected by other methods, particularly \texttt{bike} was generally below $1\mathrm{e}{-4}$, and this low noise variance led to improved predictive performance. However, Iterative GP training and predictive performance became unstable in later optimisation steps. Neither CGLB nor SGPR exhibited degradation in predictive performance during optimisation, even in cases where the noise level was near the lower bound of $1\mathrm{e}{-6}$. We hypothesise this is further evidence that lower bound maximisation is more robust as an optimisation procedure than other biased estimates of the LML, and results from optimising an objective that can overestimate the LML in ways that are non-uniform in $\theta$ (c.f.~discussion of stopping criteria in \cref{sec:cg-implementation}).

\section{Conclusion}
CGLB combines benefits from sparse variational methods with iterative approaches for hyperparameter selection. Maximising CGLB alleviates some of the hyperparameter bias that results from ELBO maximisation with sparse methods. As CGLB is deterministic, optimisation is frequently easier than using the stochastic approximations to the LML given by existing Iterative GP methods. Additionally, as the bias in CGLB results in a lower bound on the LML, we can assess choices made regarding optimisation choices using only training data, as higher lower bounds correlate well with better hyperparameter selection. 
\section*{Acknowledgements}
Thanks to Andrew Y.K.~Foong and David Janz for providing feedback on an earlier draft of this paper. DRB would like to thank the Herchel Smith Fellowship and the Qualcomm Innovation Fellowship for funding. 
\appendix 

\section{Additional bounds on the Log marginal likelihood of Gaussian process regression}

In this section, we discuss additional bounds on the log marginal likelihood of GPR regression that can be computed efficiently. We focus on the case when we have access to a rank-$m$-plus-diagonal approximate $\nQmat \prec \nKmat$. We also assume we can efficiently compute the trace of $\nKmat$. We can then think of finding the optimal lower bound on $\log p_Y(\bfy; \theta)$ as a constrained optimisation problem:
\begin{align}
    \log p_Y(\bfy; \theta) = c - \frac{1}{2}\ydat\transpose \inv{\nKmat} \ydat - \frac{1}{2}\logdet{\nKmat} \geq c+\inf_{\substack{A \succ \nQmat\\ \Tr(A)=t}}\left( -\frac{1}{2}\ydat\transpose \inv{A} \ydat - \frac{1}{2}\logdet{A}\right). 
\end{align}

We can then apply an element-wise bound, 
\begin{align}
\log p_Y(\bfy; \theta) \geq c+\inf_{\substack{A \succ \nQmat\\ \Tr(A)=t}} \left(-\frac{1}{2}\ydat\transpose \inv{A} \ydat - \frac{1}{2}\logdet{A}\right) \geq c+\inf_{\substack{A \succ \nQmat\\ \Tr(A)=t}} -\frac{1}{2}\ydat\transpose \inv{A} \ydat + \inf_{\substack{A \succ \nQmat\\ \Tr(A)=t}} - \frac{1}{2}\logdet{A} 
\end{align}
We now consider each term separately,
\subsection{Quadratic term}
Since $A\prec \nQmat$, we may write $A = \nQmat + EE$, where $E$ is the PSD square root of $A-\nQmat$. Then applying Woodbury's Lemma,
\begin{equation}
    \ydat\transpose \inv{A} \ydat = \ydat\transpose \inv{\nQmat} \ydat - \ydat\transpose\inv{\nQmat}E\inv{(I + E \inv{\nQmat}E)}E\inv{\nQmat} \ydat.
\end{equation}
The second term is non-negative, but can be $0$; in particular if $\nKmat = \nQmat + t\bfz\bfz\transpose$, where $\bfz$ is a unit vector orthogonal to $\inv{\nQmat} \ydat$. Hence, 
\begin{equation}
    \inf_{\substack{A \succ \nQmat\\ \Tr(A)=t}} -\frac{1}{2}\ydat\transpose \inv{A} \ydat = -\frac{1}{2}\ydat\transpose \inv{\nQmat} \ydat.
\end{equation}

\subsection{Log-determinant term}
We write
\begin{align}
\logdet{A} = \sum_{i=1}^n \log(a_i) = \sum  \log(\ell_i +e_i),
\end{align}
where $a_i$ are the eigenvalues of $A$, $\ell_i$ are the eigenvalues of $\nQmat$, both sorted in descending order, and $e_i:=a_i - \ell_i$. We can then translate the constraints of $\nKmat$ as $e_i\geq 0$ and $\sum_i (e_i + \ell_i) = t$. The $e_i$ are eigenvalues of $\nQmat$, which can be computed in $O(nm^2)$ by noting that $n-m$ of them are $\sigma^2$, and the remaining eigenvalues coincide with the eigenvalues of $MM\transpose + \sigma^2I$, where $M = \Kuu^{-1/2}\Kuf$. 

We define $t' = t - \Tr(\nQmat)$ and consider
\begin{align}
\sup_{\substack{A \succ \nQmat\\ \Tr(A)=t}}\logdet{A} = \sup_{\substack{e_i >0 \\ \sum e_i = t'}} \log (e_i + \ell_i).
\end{align}
This coincides with a problem in information theory related to power-allocation over channels. It is a convex optimization which can be solved using Lagrange multiplier and the KKT conditions. The solution of this equation is,
\begin{align}
    e_i = \max(0, t'/\nu - t'\ell_i),
\end{align}
where $\nu$ is chosen such that, $\sum_{i=1}^n \max(0, 1/\nu - \ell_i) = 1$. See Example 5.2 in \citet{boyd2004convex} for details of this maximization.

The eigenvalues of $\nQmat$ can be computed in $O(nm^2)$ and the Lagrange multiplier can be found in $O(n)$ as it is the solution to a piecewise linear problem.

\subsubsection{Lower bounding the log determinant}
In some cases upper bounds on the log marginal likelihood are of interest \citep{titsias_variational_2014, kim2018scaling}. We therefore turn to the problem of lower bounding the log determinant of $A$. Improvements in this bound can be used in conjunction with either of the bounds on the quadratic term given in \citet{titsias_variational_2014} or \citet{kim2018scaling}. We have,
\begin{align}
    \log(A) = \sum \log(e_i+\ell_i) = \logdet{\nQmat} + \sum \log(1 + \frac{e_i}{\ell_i}).
\end{align}
Rewriting the second term on the right hand side as the log of a product, expanding the product and using that $e_i\geq 0$, 
\begin{align}
    \sum_{i=1}^n \log\left(1 + \frac{e_i}{\ell_i}\right) = \log \prod_{i=1}^n \left(1 + \frac{e_i}{\ell_i}\right) \geq \log \left(1+ \sum_{i=1}^n \frac{e_i}{\ell_i}\right).
\end{align}
Using that $\ell_i \leq \ell_1$,
\begin{equation}
    \logdet{A} \geq \logdet{\nQmat} + \log\left(1+\frac{\Tr(\nKmat -\nQmat)}{\ell_1}\right).
\end{equation}
We now show that this is the greatest upper bound given the constraints on $A$ by constructing an $A$ satisfying this bound. Consider $A=\nQmat + \Tr(\nKmat-\nQmat)\bfw\bfw\transpose$, where $\bfw$ is the eigenvector of $\Qff$ corresponding to $\ell_1$. Then all of the eigenvalues of $\nKmat$ coincide with the eigenvalues of $\nQmat$, except that largest, which is $\ell_1+\Tr(\nKmat-\nQmat))$. Rearranging the formula for the log determinant, we see that this recovers the bound given above.

\section{Additional Experiments}

\begin{figure*}[h!]
    \centering
    \includegraphics[width=0.87\textwidth]{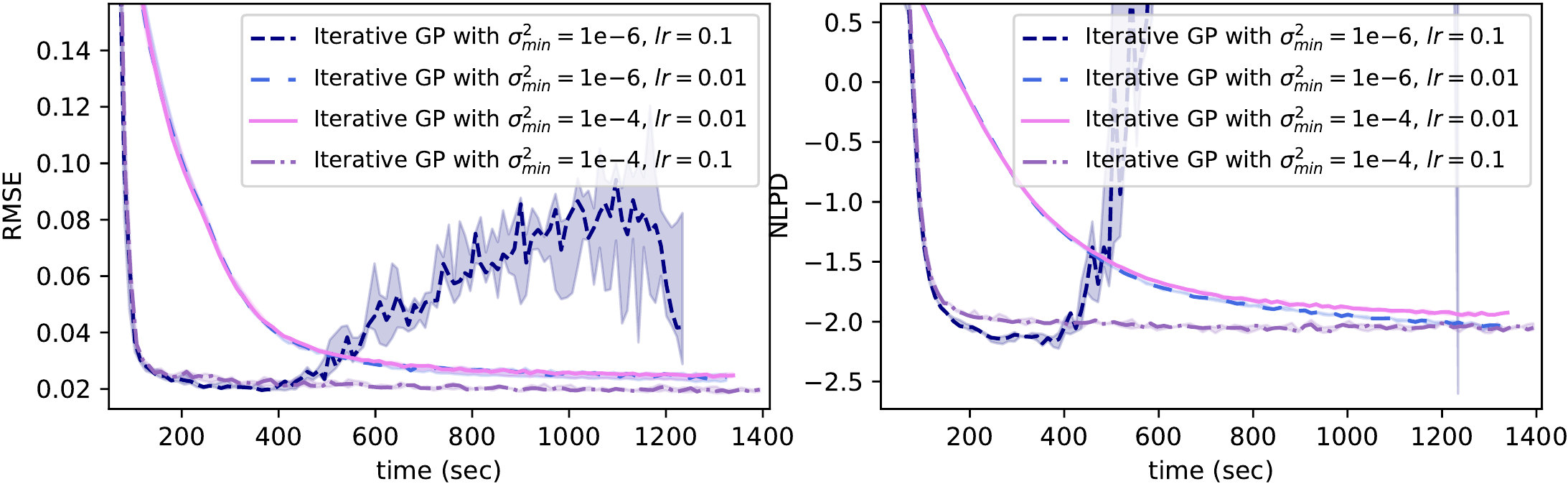}
    \caption{Test RMSE and NLPD for the Iterative GP method with learning rates $0.01$ and $0.1$ and minimum likelihood noise constrained at $1\mathrm{e}{-4}$ and $1\mathrm{e}{-6}$ on the \texttt{bike} dataset.}
    \label{fig:itergp-bike-rmse-nlpd-sup}
\end{figure*}

\begin{figure*}[h!]
    \centering
    \includegraphics[width=0.87\textwidth]{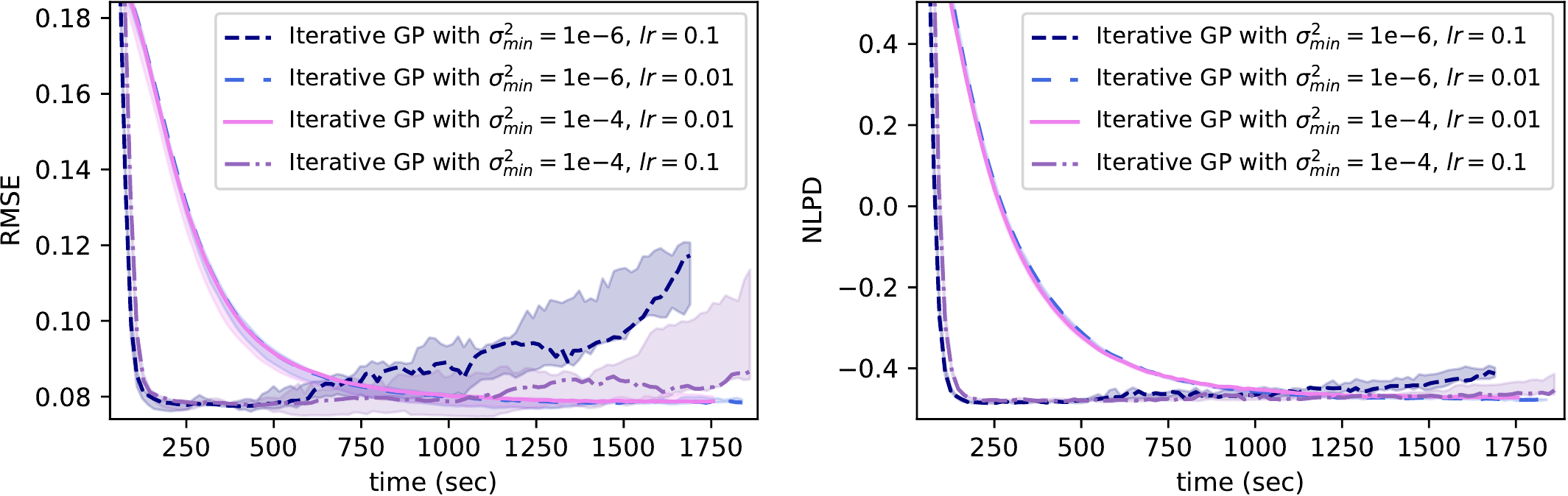}
    \caption{Test RMSE and NLPD for the Iterative GP method with learning rates $0.01$ and $0.1$ and minimum likelihood noise constrained at $1\mathrm{e}{-4}$ and $1\mathrm{e}{-6}$ on the \texttt{kin40k} dataset.}
    \label{fig:itergp-kin40k-rmse-nlpd-sup}
\end{figure*}

\begin{figure*}[h!]
    \centering
    \includegraphics[width=0.87\textwidth]{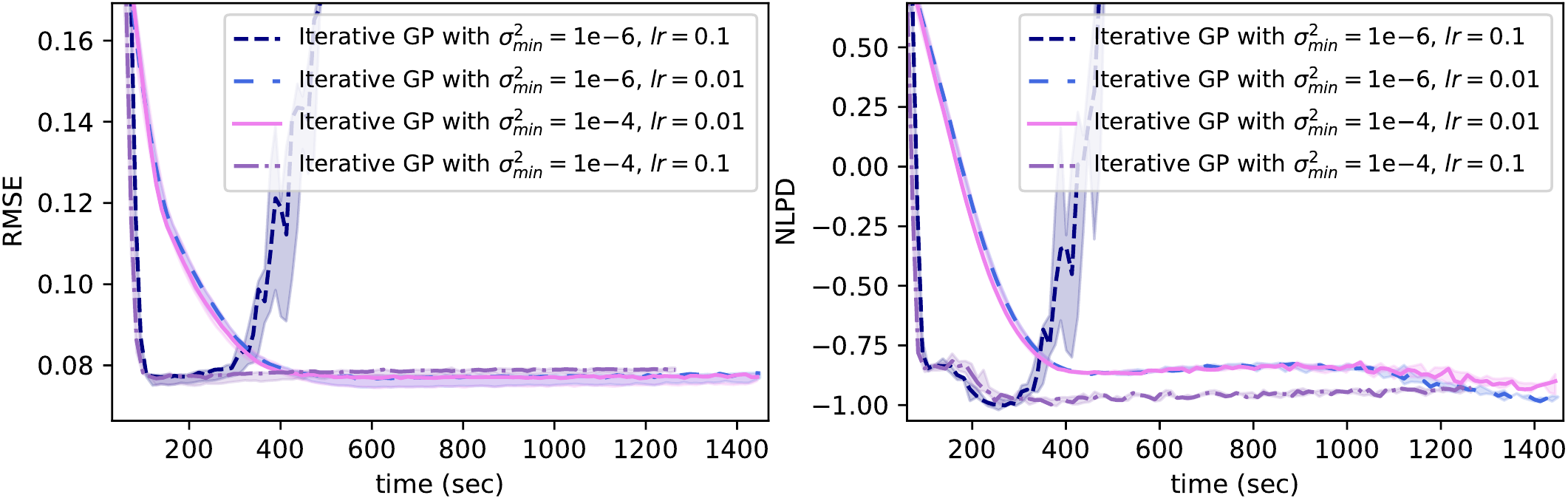}
    \caption{Test RMSE and NLPD for the Iterative GP method with learning rates $0.01$ and $0.1$ and minimum likelihood noise constrained at $1\mathrm{e}{-4}$ and $1\mathrm{e}{-6}$ on the \texttt{poletele} dataset.}
    \label{fig:itergp-pol-rmse-nlpd-sup}
\end{figure*}

\begin{figure*}[h!]
    \centering
    \includegraphics[width=0.87\textwidth]{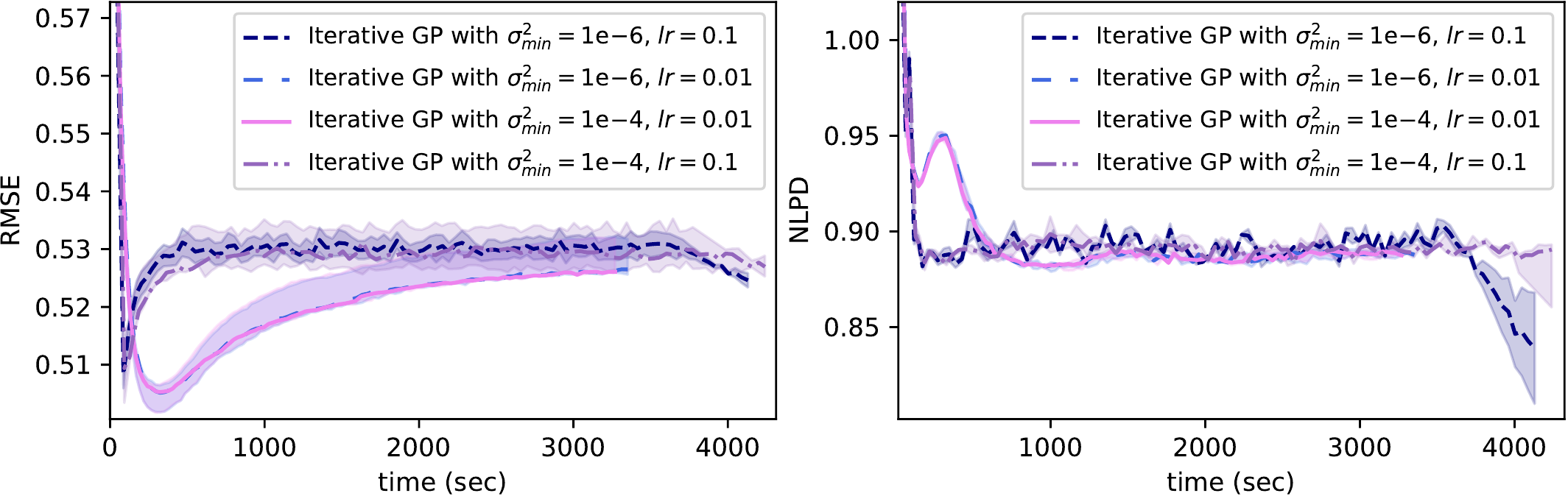}
    \caption{Test RMSE and NLPD for the Iterative GP method with learning rates $0.01$ and $0.1$ and minimum likelihood noise constrained at $1\mathrm{e}{-4}$ and $1\mathrm{e}{-6}$ on the \texttt{protein} dataset.}
    \label{fig:itergp-protein-rmse-nlpd-sup}
\end{figure*}

We performed Iterative GP experiments with all combinations of Adam learning rates ($0.1$, $0.01$) and minimum likelihood noise constraints ($\sigma^{2}_{min}=1\mathrm{e}{-4}$, $\sigma^{2}_{min}=1\mathrm{e}{-6}$) for \texttt{poletele}, \texttt{kin40k}, \texttt{elevators}, \texttt{bike} and \texttt{protein} datasets. The experiments displayed instability and sensitivity to some settings (\cref{fig:itergp-bike-rmse-nlpd-sup}, \cref{fig:itergp-kin40k-rmse-nlpd-sup}, and \cref{fig:itergp-pol-rmse-nlpd-sup}). In particular, we found that the setting with $\sigma^{2}_{min}=1\mathrm{e}{-6}$ and Adam learning rate $0.1$ (\citet{wang2019exact} uses $\sigma^{2}_{min}=1\mathrm{e}{-4}$ and Adam learning rate $0.1$), causes severe predictive performance degradation in \texttt{poletele}, \texttt{bike}, and \texttt{kin40k} datasets after $2000$ iterations. Lowering the learning rate to $0.01$ facilitates smoother, but much slower convergence. However, datasets such as \texttt{poletele} (\cref{fig:itergp-pol-rmse-nlpd-sup}) and \texttt{protein} (\cref{fig:itergp-protein-rmse-nlpd-sup}) still exhibit a decrease in the predictive performance during training.

\begin{figure*}[h!]
    \centering
    \includegraphics[width=0.87\textwidth]{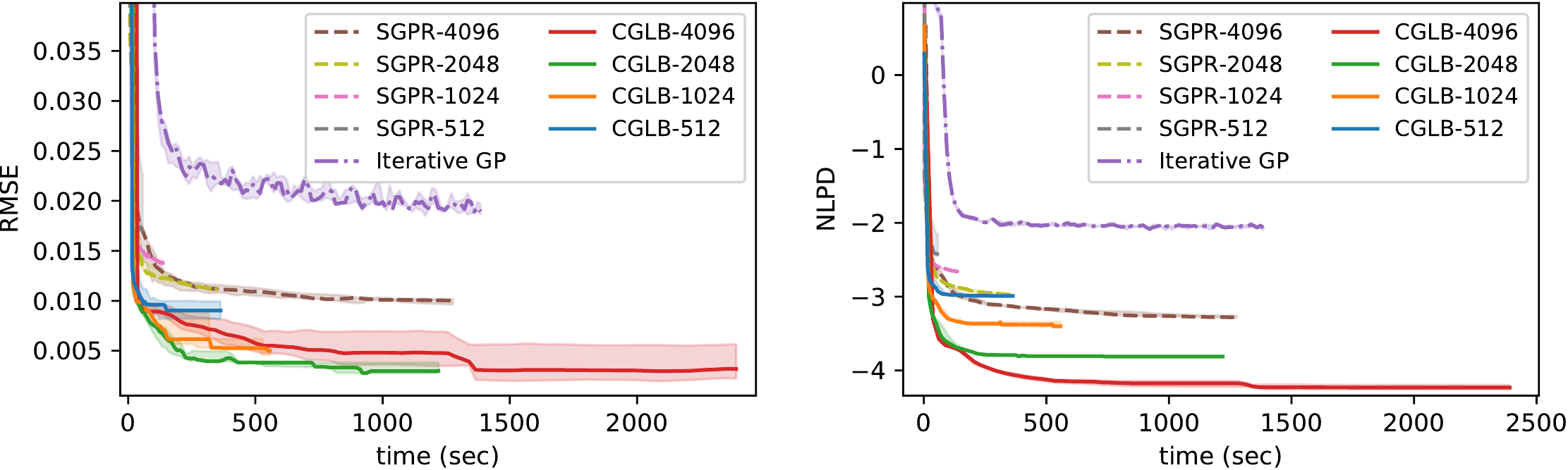}
    \caption{Test root mean square error (RMSE) and negative log predictive density (NLPD) metrics of CGGP, SGPR and Iterative GP models computed on \texttt{bike} dataset. The shaded area is IQR region, and the line is a median over five experiment trials with different dataset splits.}
    \label{fig:bike-rmse-nlpd-sup}
\end{figure*}

\begin{figure*}[h!]
    \centering
    \includegraphics[width=0.87\textwidth]{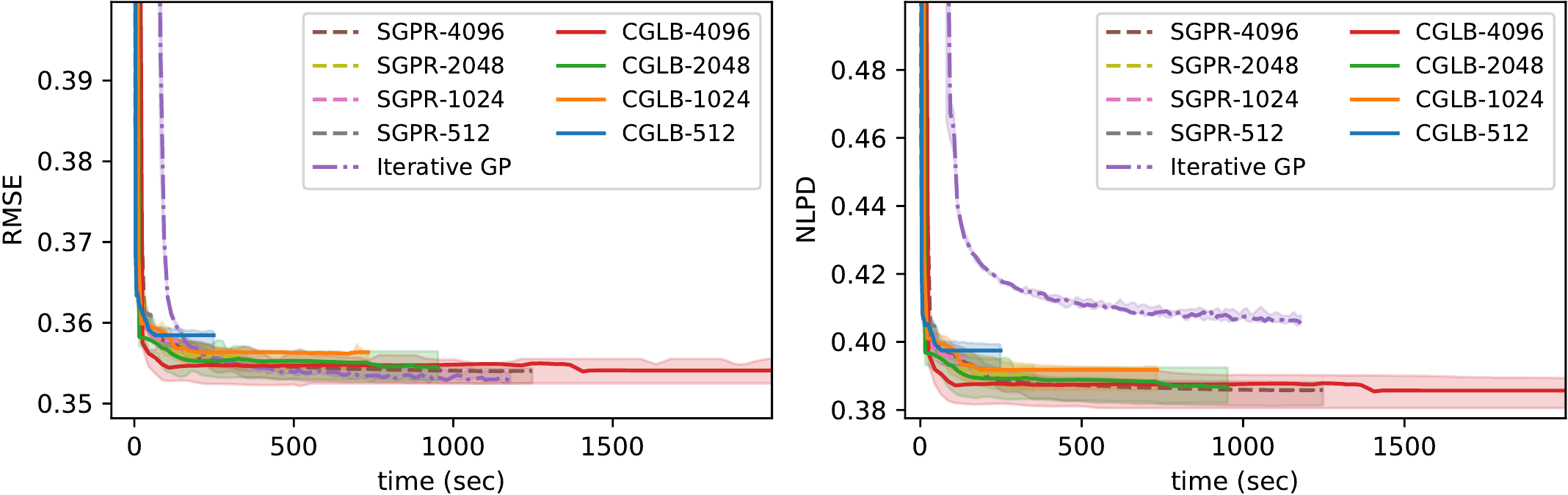}
    \caption{Test root mean square error (RMSE) and negative log predictive density (NLPD) metrics of CGGP, SGPR and Iterative GP models computed on \texttt{elevators} dataset. The shaded area is IQR region, and the line is a median over five experiment trials with different dataset splits.}
    \label{fig:elevators-rmse-nlpd-sup}
\end{figure*}

\begin{figure*}[h!]
    \centering
    \includegraphics[width=0.87\textwidth]{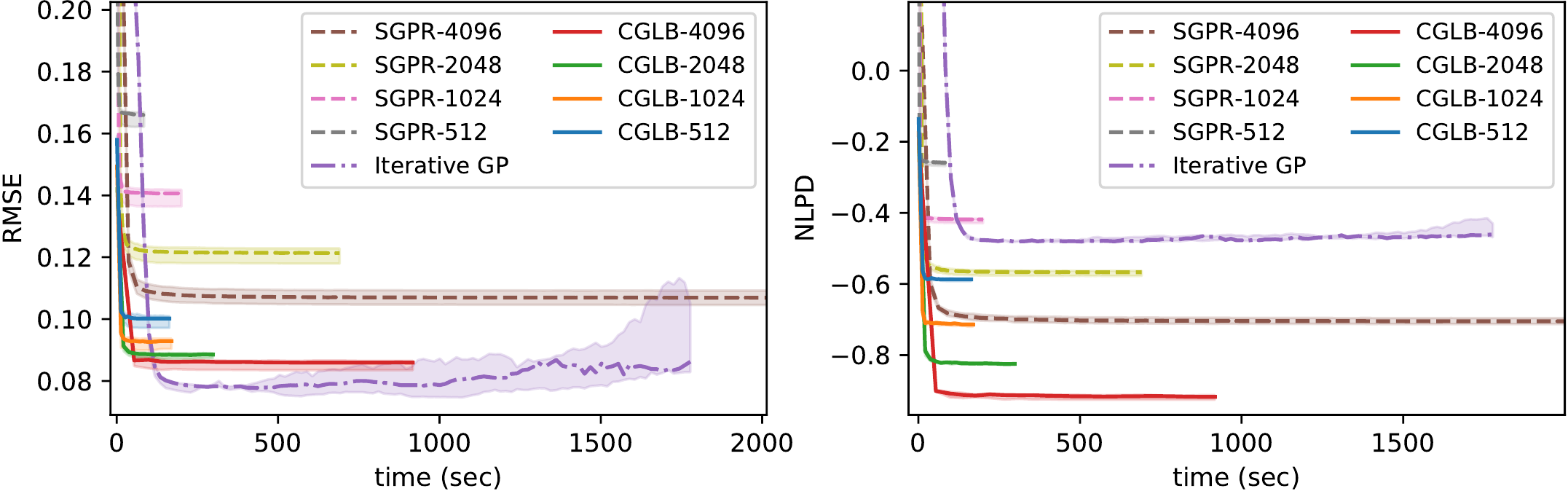}
    \caption{Test root mean square error (RMSE) and negative log predictive density (NLPD) metrics of CGGP, SGPR and Iterative GP models computed on \texttt{kin40k} dataset. The shaded area is IQR region, and the line is a median over five experiment trials with different dataset splits.}
    \label{fig:kin40k-rmse-nlpd-sup}
\end{figure*}

\begin{figure*}[h!]
    \centering
    \includegraphics[width=0.87\textwidth]{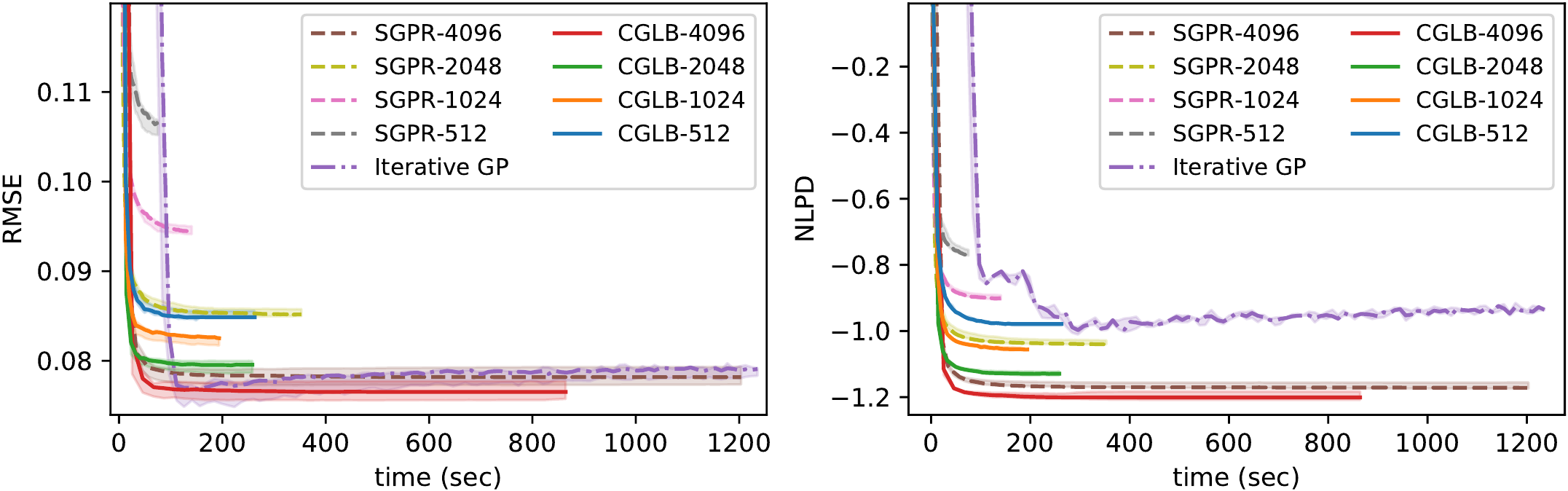}
    \caption{Test root mean square error (RMSE) and negative log predictive density (NLPD) metrics of CGGP, SGPR and Iterative GP models computed on \texttt{poletele} dataset. The shaded area is IQR region, and the line is a median over five experiment trials with different dataset splits.}
    \label{fig:pol-rmse-nlpd-sup}
\end{figure*}

\begin{figure*}[h!]
    \centering
    \includegraphics[width=0.87\textwidth]{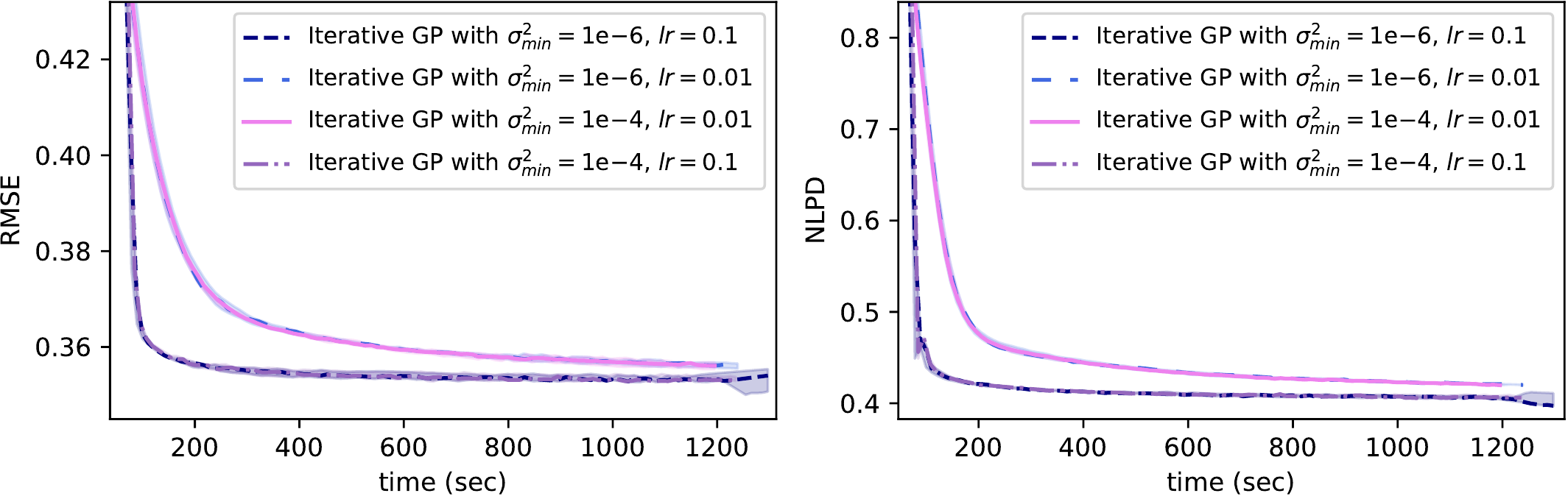}
    \caption{Test RMSE and NLPD for the Iterative GP method with learning rates $0.01$ and $0.1$ and minimum likelihood noise constrained at $1\mathrm{e}{-4}$ and $1\mathrm{e}{-6}$ on the \texttt{elevators} dataset.}
    \label{fig:itergp-elevators-rmse-nlpd-sup}
\end{figure*}

\clearpage
\bibliography{main}
\bibliographystyle{plainnat}
\end{document}